\documentclass[letterpaper, 10 pt, conference]{ieeeconf}  % Comment this line out if you need a4paper

\IEEEoverridecommandlockouts                              % This command is only needed if 
\usepackage{graphicx}
\usepackage{booktabs}
\usepackage{multirow}
\usepackage{hhline}
\usepackage{siunitx}
\usepackage{algorithm, algorithmic}

\usepackage{amsmath,amsfonts}
\usepackage{array}
\usepackage[caption=false,font=footnotesize,labelfont=rm,textfont=rm]{subfig}
\usepackage{textcomp}
\usepackage{stfloats}
\usepackage{url}
\usepackage{verbatim}
\usepackage{graphicx}
\usepackage{cite}
\usepackage{times}
\usepackage{epsfig}
\usepackage{float}
\usepackage{wrapfig}
\usepackage{algorithm, algorithmic}
\usepackage{bm,xspace}
\usepackage{comment}
\usepackage{multirow}
\usepackage{threeparttable}
\usepackage{balance}
\usepackage{booktabs}
\usepackage{etoolbox,siunitx}
\usepackage{calc}
\usepackage{pifont,hologo}
\usepackage{makecell}
\usepackage{adjustbox}
\usepackage{float}
\usepackage[colorlinks=black,citecolor=green]{hyperref}

\makeatletter
\newenvironment{breakablealgorithm}
{% \begin{breakablealgorithm}
		\begin{center}
			\refstepcounter{algorithm}% New algorithm
			\hrule height.8pt depth0pt \kern2pt% \@fs@pre for \@fs@ruled
			\renewcommand{\caption}[2][\relax]{% Make a new \caption
				{\raggedright\textbf{\ALG@name~\thealgorithm} ##2\par}%
				\ifx\relax##1\relax % #1 is \relax
				\addcontentsline{loa}{algorithm}{\protect\numberline{\thealgorithm}##2}%
				\else % #1 is not \relax
				\addcontentsline{loa}{algorithm}{\protect\numberline{\thealgorithm}##1}%
				\fi
				\kern2pt\hrule\kern2pt
			}
		}{% \end{breakablealgorithm}
		\kern2pt\hrule\relax% \@fs@post for \@fs@ruled
	\end{center}
}
\makeatother

\title{\LARGE \bf
Design and Control of an Ultra-Slender Push-Pull Multisection Continuum Manipulator for In-Situ Inspection of Aeroengine
}

\author{Weiheng Zhong$^{1}$, Yuancan Huang$^{1}$, \textit{Senior Member}, \textit{IEEE}, Da Hong$^{1}$, Nianfeng Shao$^{1}$% <-this % stops a space
\thanks{$^{1}$Weiheng Zhong, Yuancan Huang, Da Hong, and Nianfeng Shao are with the School of Mechatronical Engineering, Beijing Institute of Technology, 100081, China
        {\tt\small \{weiheng\_z, yuancanhuang, hongda, shaonianfeng95\}@bit.edu.cn}}%
}

\begin{document}

\maketitle
\thispagestyle{empty}
\pagestyle{empty}

\begin{abstract}

Since the shape of industrial endoscopes is passively altered according to the contact around it, manual inspection approaches of aeroengines through the inspection ports have unreachable areas, and it's difficult to traverse multistage blades and inspect them simultaneously, which requires engine disassembly or the cooperation of multiple operators, resulting in efficiency decline and increased costs.
To this end, this paper proposes a novel continuum manipulator with push-pull multisection structure which provides a potential solution for the disadvantages mentioned above due to its higher flexibility, passability, and controllability in confined spaces.
The ultra-slender design combined with a tendon-driven mechanism makes the manipulator acquire enough workspace and more flexible postures while maintaining a light weight. 
Considering the coupling between the tendons in multisection, a innovative kinematics decoupling control method is implemented, which can realize real-time control in the case of limited computational resources.
A prototype is built to validate the capabilities of mechatronic design and the performance of the control algorithm.
The experimental results demonstrate the advantages of our continuum manipulator in the in-situ inspection of aeroengines' multistage blades, which has the potential to be a replacement solution for industrial endoscopes.

\end{abstract}

\begin{figure}[!h]
    \centering
    \includegraphics[width=0.95\linewidth]{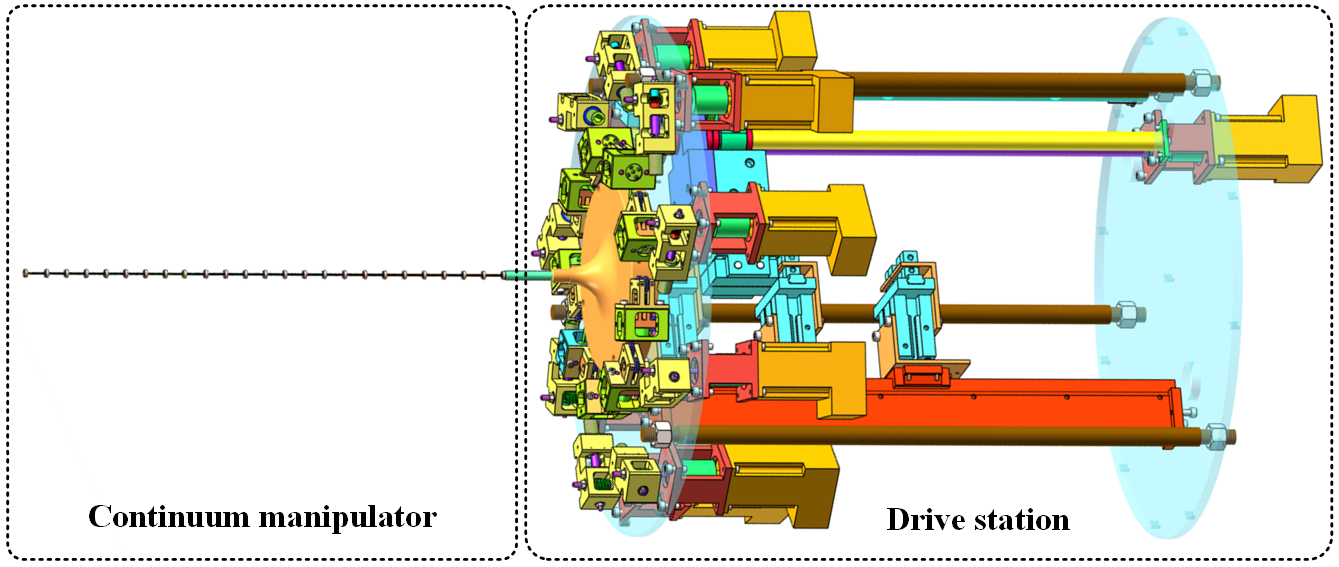}
    \caption{Conceptual illustration of the PPCM.}
    \label{fig:illustration}
\end{figure}

\section{INTRODUCTION}

Industrial endoscope-assisted manual in-situ inspection of the areoengine through the inspection ports reserved in its casing is the mainstream approach at present.
However, traversing the multistage blades with the endoscope proves challenging due to its passive alteration in shape upon contact with surroundings, which necessitates engine disassembly and the collaborative effort of inspectors to maneuver the detection head into invisible areas for comprehensive blade inspection, resulting in efficiency decline and increased costs.

\begin{figure}[!tb]
    \centering
    \includegraphics[width=0.95\linewidth]{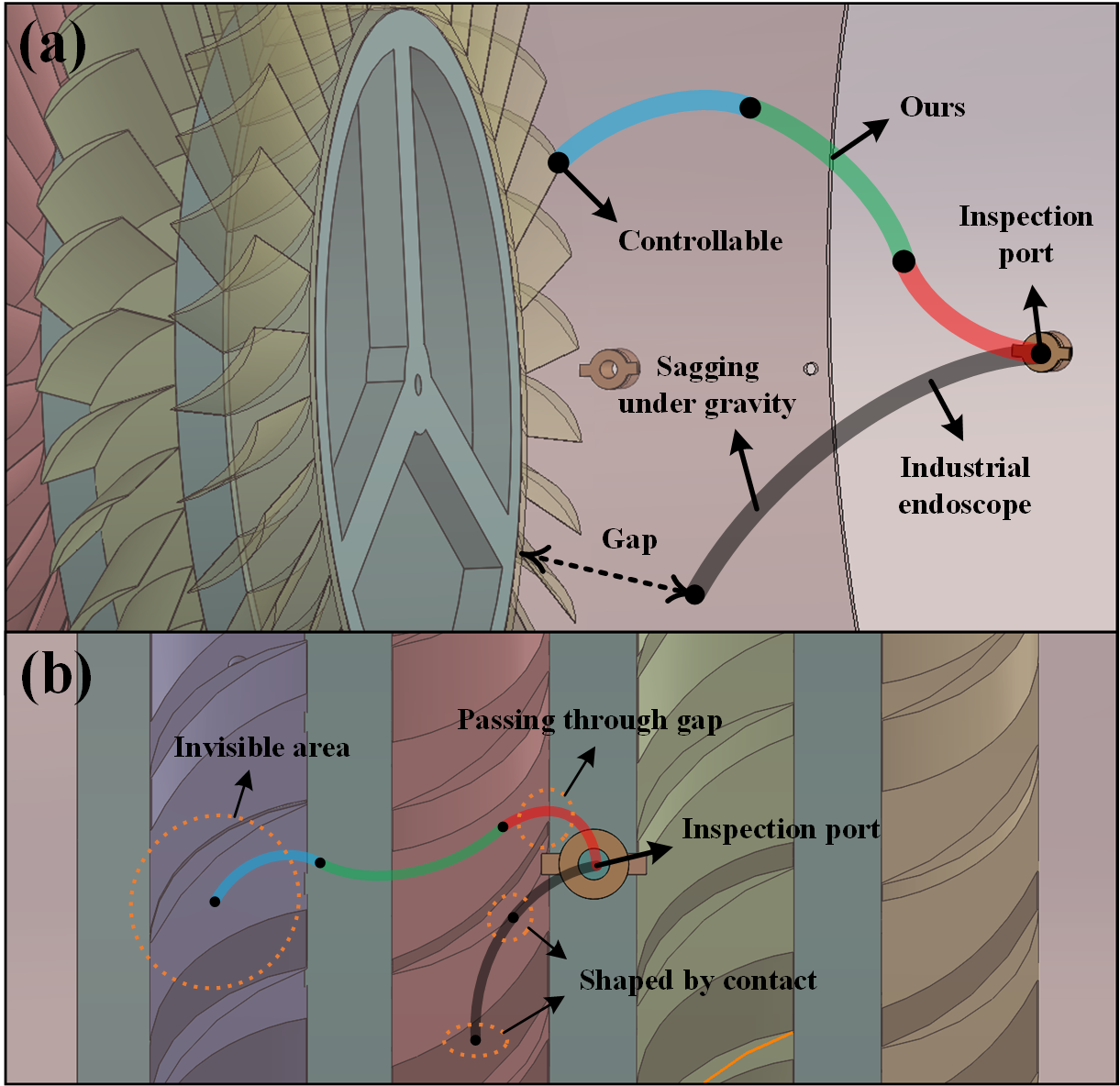}
    \caption{a) Turbine inspection scenario; b) Compressor inspection scenario.}
    \label{fig:comparison}
\end{figure}

Due to the mature and extensive application of continuum manipulators in medicine, researchers are increasingly interested in their potential capabilities for industrial inspection \cite{continuumreview}.
Several teams have previously explored or implemented research focusing primarily on mechanism design (including kinematics) \cite{aeroengine1, aeroengine2, otherapplication, limited}, control methods \cite{fuzzy}, and the development of various working tools \cite{hightemp, limited} or inspection sensors \cite{nanhang} for continuum manipulators.
The research in this paper centers around mechanical design and control implementation.
% Several teams have already discussed this research or implemented it in applications, which mainly focus on the mechanism design (include kinematics) \cite{aeroengine1}, \cite{aeroengine2}, \cite{otherapplication}, \cite{limited}, control methods \cite{fuzzy}, and the development of various working tools \cite{hightemp}, \cite{limited} or inspection sensors \cite{nanhang} for the continuum manipulator, and the research of this paper revolves around mechanical design and control implementation.

Based on a comparative analysis of existing achievements in practical applications, we have identified several key areas that require further improvement.
The ultra-slender structure developed in \cite{aeroengine2} successfully overcomes the working range limitations of continuum manipulators, allowing for application in a turbine inspection scenario. However, the lack of push-pull functionality restricts its ability to traverse narrow gaps, and thus failing to meet the challenge of inspecting multiple stages of blades through a single inspection port in a compressor scenario.
% Moreover, it does not provide the versatility required for diverse inspection scenarios.
% The work in \cite{aeroengine2} breaks through the limitation of the continuum manipulator's working range.
% However, the ultra-slender structure developed in the paper may only be suitable for single-stage blades, as it lacks the capability for push-pull functionality.
% This structure does not address the challenges of multistage blade inspection under a single inspection port and the versatility required for multiple scenarios.
While \cite{otherapplication} and \cite{nanhang} have basically achieved the traversal of multi-stage blades, their proposed mechanical designs lack consideration for constraints such as inspection port limitations, blade stagger angle, and spacing between blades, making them unsuitable for more challenging real-world scenarios.
% On the contrary, while the achievement of traversing multistage blades is demonstrated in \cite{otherapplication} and \cite{nanhang}, the mechanical design presented lacks addressing the challenges posed by the limitations of the inspection port, blade stagger angle and spacing between blades, thus falling short of meeting practical requirements.

To address the aforementioned shortcomings, we developed an ultra-slender push-pull multisection continuum manipulator (PPCM) with a length-to-diameter ratio of up to 62.5 (56.3 as reported in \cite{aeroengine2}), as illustrated in Fig. \ref{fig:illustration}, which, to our knowledge, represents the most slender design for continuum robots in aeroengine inspection.
The PPCM facilitates multistage blade inspection through more intuitive teleoperation and has been validated for its versatility in various aeroengine inspection scenarios using a small simulation platform built from real data, which presents greater challenges compared to larger aeroengines.
% The push-pull multisection structure enables the continuum manipulator to achieve multistage blade inspection through a more intuitive teleoperation method, eliminating the need for complex motion planning.
% Based on these designs, we implement a decoupling control for a tendon-driven mechanism.
% Furthermore, we validated the universality of the PPCM in various aeroengine inspection scenarios using a small simulation platform constructed with real data, which posed greater challenges compared to larger aeroengines.

In summary, our contributions are as follows:
\begin{itemize}
\item An ultra-slender continuum manipulator with push-pull multisection structure is developed.
\item A kinematic decoupling real-time control method for tendon-driven mechanism is implemented.
\item The advantages and adaptability of our continuum manipulator in various in-situ inspection scenarios are validated in a simulation platform for a certain type of aeroengine.
\end{itemize}

\begin{figure*}[!t]
    \centering
    \includegraphics[width=0.85\textwidth]{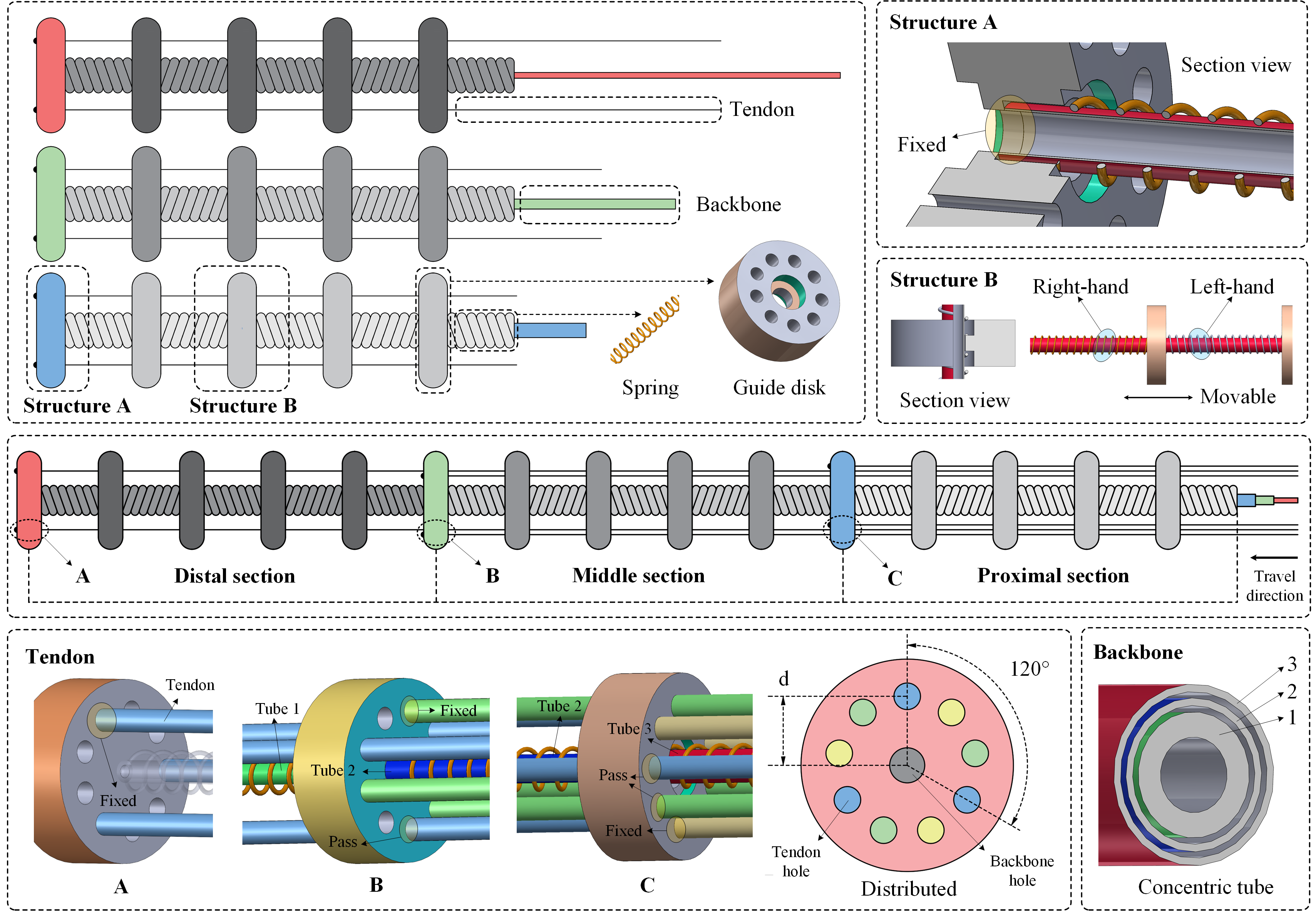}
    \caption{Schematic of the continuum manipulator: It is assembled by three single sections, of which three backbones form a concentric tube structure, and tendons between the three sections are coupled to each other.}
    \label{fig:continuum manipulator structure}
\end{figure*}

\section{MECHANICAL DESIGN}

\subsection{Requirements Identification}

Specific inspection scenarios of the aeroengine include compressors with multiple stages of blades and high- and low-pressure turbines.
The inspection port of the turbine is situated at a distance from the turbine, as shown in Fig. \ref{fig:comparison}(a), and there are no contact points within the space for the endoscope to properly align, resulting in sagging under gravity.
The inspection port of the compressor is located in the middle of the two-stage blades, as shown in Fig. \ref{fig:comparison}(b). The confined space between the two-stage blades, along with the large blade stagger angle, poses challenges for the endoscope to reach deeper locations.
% The design requirements of the PPCM are determined by the aeroengine's inspection scenarios, which consist of multistage compressors and high- and low-pressure turbines.
% The inspection port for the compressor situated on the engine casing, positioned in the middle of the entire compressor blades, is constrained in use by the limited space between the port and blades, and the larger blade stagger angle further diminishes the available space between the blades.
% The distance between the inspection port near the aeroengine's tail nozzle and the turbine is considerable; moreover, the space between the two is unobstructed, signifying that there are no components between them.

The difficulties presented by the aforementioned two inspection scenarios are to be addressed in the design of the PPCM, and several constraints must be taken into consideration \cite{aeroengine1}.
The various diameters and orientations (horizontal or obliquely upward) of the inspection ports determine the maximum diameter of the PPCM and the adjustability of the position and orientation of the detection probe.
The working length range of the PPCM is crucial for allowing it to maneuver freely in confined spaces as well as traverse large distances in open areas.
It is necessary to consider the maximum length and curvature of each section of the PPCM to improve its adaptability to varying blade spacing and stagger angles during the traversal of multiple stages of blades.
Moreover, the appropriate minimum payload capacity of the PPCM is essential to ensuring the successful integration of the inspection sensor.
% To enable the PPCM to meet the requirements of the aforementioned inspection scenarios, several design constraints must be considered \cite{aeroengine1}. 
% Considering various inner diameters and orientations (horizontal or oblique upward) of the inspection ports, determining the appropriate maximum diameter and adjustable orientation feature is the first step.
% After the PPCM enters the inspection port, the range of working length is critical to enable the PPCM to move freely in the compressor scenario where movement space is limited and reach the blade to be inspected in the turbine scenario where a wide range of movement is required. 
% In the process of traversing multiple blades, the maximum length and curvature of a single section determine the PPCM's adaptability at various intervals and stagger angles.
% Last but not least, the appropriate minimum payload capacity of the PPCM's tip is fundamental to ensuring the successful equipment of the inspection sensor.

\subsection{Structure Analysis}

\begin{figure}[!b]
    \centering
    \includegraphics[width=0.95\linewidth]{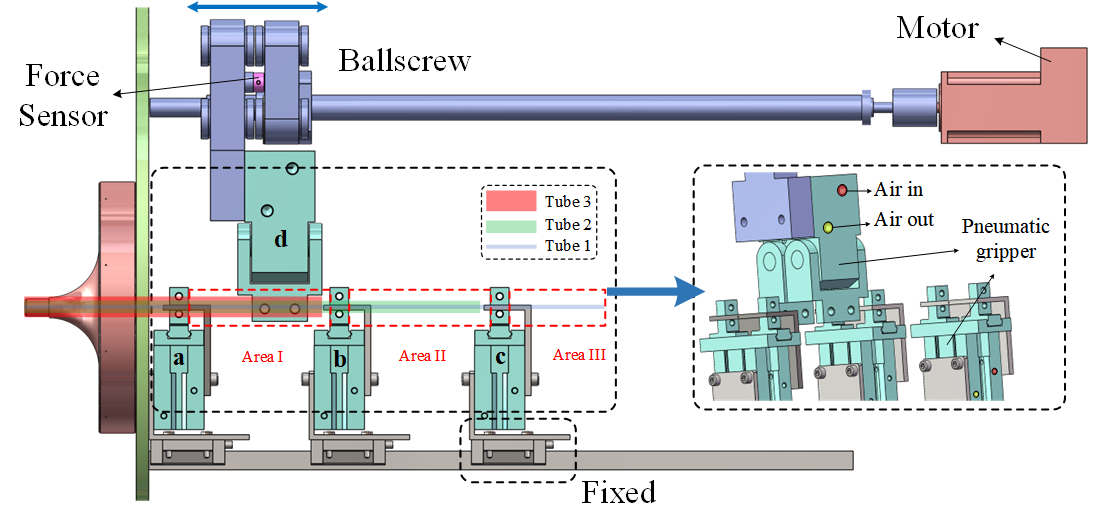}
    \caption{Schematic of the backbone actuation: It features a pneumatic system comprising three fixed pneumatic grippers and one movable gripper.}
    \label{fig:backbone actuation}
\end{figure}

% The proposed PPCM adopts extrinsic actuation mechanisms, in which motion is transferred along the manipulator by mechanical components and actuators are positioned far from the manipulator body, resulting in a continuum manipulator and its drive station, which acts as the continuum manipulator's motion generator.
The extrinsic actuation mechanism of the PPCM is fundamental to its application in aeroengine inspection scenarios, where the motion of the continuum manipulator is converted into motion at the drive station through mechanical components.
Placing the drive station far from the continuum manipulator makes it possible to miniaturize the manipulator to fit tight spaces.
In addition, considering that the continuum manipulator needs to better handle motion within irregular spaces, a three-section design is not only necessary but also reasonable.
Although the dexterity of the continuum manipulator is directly related to the number of sections, this requires being balanced with the challenges associated with achieving more precise control.

% The flexibility of the continuum manipulator is proportional to the number of sections, and having too many sections will increase the structure complexity and diameter of the continuum manipulator, which makes the implementation of control difficult.

% The flexibility of the continuum manipulator is directly related to the number of sections, and increasing the number of sections results in greater structural complexity and diameter, posing challenges for control implementation. 
% Therefore, a three-section design is developed, as shown in Fig. \ref{fig:continuum manipulator structure}, which also takes into account the tradeoff between the continuum manipulator's dexterity and the precision of motion control.

The tendon-driven mechanism represents the optimal method for achieving a highly slender configuration in continuum manipulators, thereby reducing the manipulator's diameter to accommodate additional inspection ports of varying diameters and blades with different gaps.
As illustrated in Fig. \ref{fig:continuum manipulator structure}, each section consists of a flexible tube with several guide disks with evenly spaced holes as the backbone.
The guide disk located at the end of the backbone is fixed to provide leverage for tendons to alter the shape of the backbone (see Structure A), while the other guide disks slide along the axial direction of the backbone.
The spacing between the guide disks is maintained by springs, alternating between left- and right-handed springs effectively reduces the torsional effects on the slender rod (see Structure B).
% Each section consists of a highly elastic hollow flexible tube as the central backbone, surrounded by several concentric circular guide disks with evenly spaced holes.
% To actuate the backbone, a tendon-driven structure is adopted, which effectively reduce the manipulator's overall weight and mitigates the need for compensating for self-weight.
% Also, the diameter of the continuum manipulator is well reduced to accommodate more inspection ports with various diameters and blades with various gaps.
% Given the heightened torsional deformation experienced by the slender backbone, the guide disk positioned at the tip is fixed.
% Furthermore, all guide disks are interconnected alternately using left- and right-hand springs to effectively mitigate torsional effects.

Intuitively, stacking hollow tubes of varying diameters to form a concentric tube structure enables push-pull functionality, allowing the continuum manipulator to achieve significant length variation and adapt to inspection requirements at various depths.
Compared to fixed-length structures of other continuum manipulators, this enhances the flexibility and maneuverability in multistage blade scenarios, significantly reducing the difficulty of motion planning in confined spaces.
Although the shape of a single section is controlled by three tendons, the shape of the distal, middle, and proximal sections is influenced by three, six, and nine tendons, respectively, as tendons far from the drive station traverse all guide disks of the section near the drive station (refer to the bottom left of Fig. \ref{fig:continuum manipulator structure}).
Each tendon is individually actuated and contains a tension-measuring unit.
As illustrated in Fig. \ref{fig:backbone actuation}, the fixation and release of the backbones of the three sections are controlled by pneumatic grippers 1, 2, and 3, respectively. Pneumatic gripper 4 is mounted on a ballscrew mechanism controlled by a motor, and its position in different areas can be combined with other pneumatic grippers to achieve pushing and pulling motion of a single backbone.
Table \ref{tab:status} displays the control statuses for backbone actuation, with any other combination of statuses not listed being considered invalid.

\begin{table}[!t]
    \centering
    \caption{control status of the backbone actuation}
    \label{tab:status}
    \begin{threeparttable}
    \begin{tabular}{ccccccc} 
        \toprule
        \multicolumn{4}{c}{Pneumatic gripper} &\multicolumn{3}{c}{Tube status\tnote{\dag}} \\
        \cmidrule(l){1-4}
        \cmidrule(l){5-7}
        % Gripper\ a &Gripper\ b &Gripper\ c &Gripper\ d &Tube\ 1 &Tube\ 2 &Tube\ 3 \\
        a\tnote{*} &b\tnote{*} &c\tnote{*} &d\tnote{$\star$} &1 &2 &3 \\
        \midrule
        % Open &Open &Open &Area\ I &Controllable &Movable &Movable \\
        $\checkmark$ &$\checkmark$ &$\checkmark$ &I &Controllable &Movable &Movable \\
        % Close &Open &Open &Area\ II &Fixed &Controllable &Movable \\
        $\times$ &$\checkmark$ &$\checkmark$ &II &Fixed &Controllable &Movable \\
        % Close &Close &Open &Area\ III &Fixed &Fixed &Controllable \\
        $\times$ &$\times$ &$\checkmark$ &III &Fixed &Fixed &Controllable \\
        % Close &Close &Close &Any &Fixed &Fixed &Fixed \\
        $\times$ &$\times$ &$\times$ &Any &Fixed &Fixed &Fixed \\
        \bottomrule
    \end{tabular}
    
    \begin{tablenotes}
        \footnotesize
        \item[*] The states of grippers a, b, and c: open ($\checkmark$) and close ($\times$).
        \item[$\star$] The position of gripper d: area I, II, and III. 'Any' represents that the gripper can be in any of the three areas.
        \item[\dag] 'Controllable' means capable of active movement by the gripper d. 'Movable' means passive movement. 'Fixed' means unable to move.
    \end{tablenotes}
    \end{threeparttable}
\end{table}

\subsection{Prototype Implementation}

\begin{table*}[t]
    \centering
    \caption{Structure parameters of the continuum manipulator}
    \renewcommand\arraystretch{1.5}
    \resizebox{1.0\linewidth}{!}
    {
    \begin{threeparttable}
        \begin{tabular}{lccccccccccc}
            \toprule
            \multirow{2}{*}{\begin{tabular}[c]{@{}c@{}}$\textrm{Parameters}$\end{tabular}} &\multicolumn{5}{c}{Backbone} &\multicolumn{6}{c}{Guide disk} \\
            \cmidrule(l){2-6}
            \cmidrule(l){7-12}
            
            & $\phi_\textrm{inner}\ \textrm{(mm)}$ & $\phi_\textrm{outer}\ \textrm{(mm)}$ & \textrm{length (mm)} & \textrm{mass (g)} & material & $\phi_\textrm{tube\ hole}\ \textrm{(mm)}$ & mass (g) & $\phi_\textrm{tendon\ hole}\ \textrm{(mm)}$ & $\phi_\textrm{outer}\ \textrm{(mm)}$ & height\ \textrm{(mm)} & material \\
            
            \midrule
            $\textrm{Proximal\ section}$ & 1.32 & 1.44 & 380 & 1.14 & \multirow{3}{*}{Ni-Ti} & 1.59 & 0.19 & \multirow{3}{*}{1$\pm$0.1} & \multirow{3}{*}{8$\pm$0.05} & \multirow{3}{*}{3$\pm$0.1} & \multirow{3}{*}{Resin} \\
            $\textrm{Middle\ section}$ & 1.05 & 1.17 & 678 & 1.20 & & 1.32 & 0.17 & & & & \\
            $\textrm{Distal\ section}$ & 0.56 & 0.9 & 975 & 1.85 & & 1.05 & 0.14 & & & & \\
            
            \midrule
            \multirow{2}{*}{Tendon} & \multicolumn{2}{c}{$\phi\ \textrm{(mm)}$} & \multicolumn{4}{c}{material} & \multicolumn{2}{c}{plied\ yarn \textrm{(Number)}} & \multicolumn{3}{c}{yarn\ count\ \textrm{(Nden)}} \\
            & \multicolumn{2}{c}{0.3} & \multicolumn{4}{c}{ultra-high molecular weight polyethylene fibers} & \multicolumn{2}{c}{3} & \multicolumn{3}{c}{200} \\
            
            \toprule
            {Parameters} & \multicolumn{2}{c}{$l_\textrm{min}\ \textrm{(mm)}$} & \multicolumn{2}{c}{$l_\textrm{max}\ \textrm{(mm)}$} & \multicolumn{3}{c}{length-to-diameter\ ratio} & \multicolumn{2}{c}{bending\ capability} & \multicolumn{2}{c}{mass\ \textrm{(g)}} \\
            
            \midrule
            Continuum manipulator & \multicolumn{2}{c}{160} & \multicolumn{2}{c}{502} & \multicolumn{3}{c}{20 $\sim$ 62.75} & \multicolumn{2}{c}{Over $180^\circ$} & \multicolumn{2}{c}{13.376} \\
            
            Proximal\ section & \multicolumn{2}{c}{38} & \multicolumn{2}{c}{162} & \multicolumn{3}{c}{4.75 $\sim$ 20.25} & \multicolumn{2}{c}{About $75^\circ$} & \multicolumn{2}{c}{3.976} \\
            Middle\ section & \multicolumn{2}{c}{44} & \multicolumn{2}{c}{158} & \multicolumn{3}{c}{5.50 $\sim$ 19.75} & \multicolumn{2}{c}{About $75^\circ$} & \multicolumn{2}{c}{3.960} \\
            Distal\ section & \multicolumn{2}{c}{78} & \multicolumn{2}{c}{182} & \multicolumn{3}{c}{9.75 $\sim$ 22.75} & \multicolumn{2}{c}{About $85^\circ$} & \multicolumn{2}{c}{5.440} \\
            
            \bottomrule
            
        \end{tabular}
        \label{tab:parameter}
    \end{threeparttable}
    }
\end{table*}

\begin{figure}[!b]
    \centering
    \includegraphics[width=0.95\linewidth]{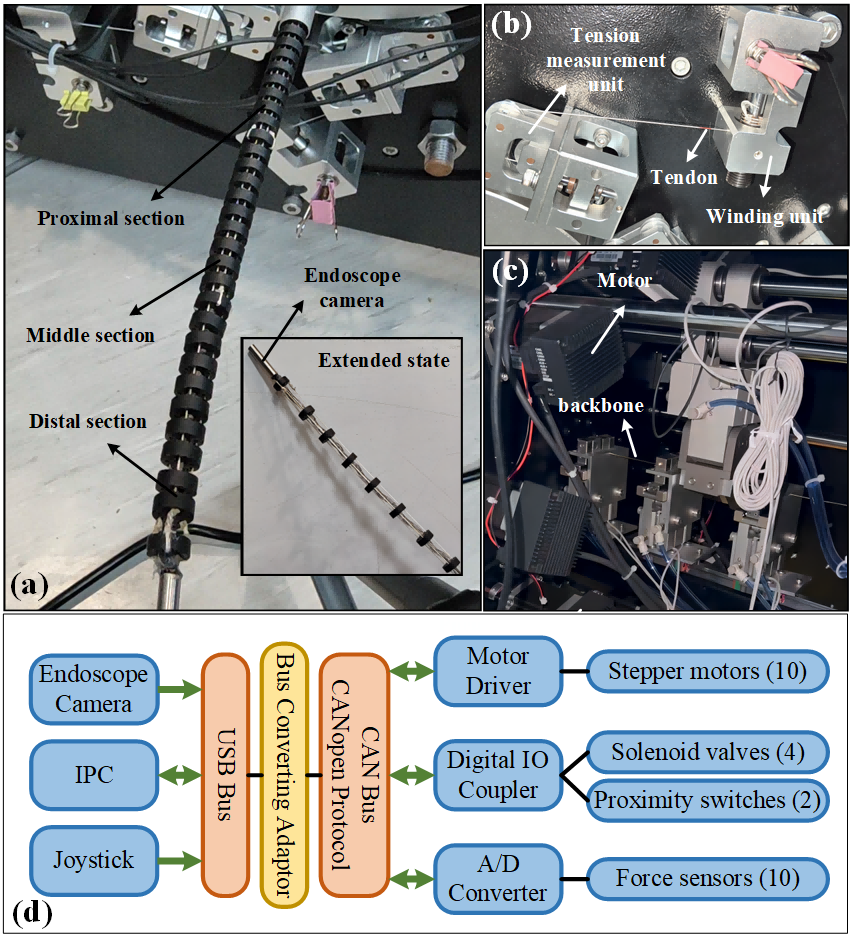}
    \caption{Prototype of the PPCM. a) Continuum manipulator; b) A set of tendon actuation units comprising a winding unit and a force measurement unit; c) Backbone actuation; d) Architecture of the mechatronic system.}
    \label{fig:implementation}
\end{figure}

A prototype of the PPCM is implemented as depicted in Fig. \ref{fig:implementation}, with the structural parameters detailed in Table \ref{tab:parameter}.
Using Ni-Ti hollow tubes as the backbone of the continuum manipulator ensures its bending resistance (preventing excessive sagging under gravity) and high toughness (recovery from large-angle bending), while maintaining a certain level of lightweight design.
Furthermore, guide disks made of resin material using a light-curing printer further reduce weight compared to metal materials (\cite{torontoiros} and \cite{toronto}), which are crucial for enhancing the length-to-diameter ratio of the continuum manipulator and mitigating the need for compensating for self-weight.
Correspondingly, tendons must possess excellent tensile and shear strength to provide significant torque to the backbone. Therefore, utilizing braided wire made from ultra-high-molecular-weight polyethylene fiber (five times that of nylon) with outstanding wear resistance is an intuitive choice for tendons. In addition, the torque of the stepper motor actuating the tendon reaches 1.5 \si{\newton\cdot\meter}.
To ensure the effectiveness of pneumatic grippers in controlling the deformation and axial movement of the backbone, the clamping force of the grippers is maintained by the pneumatic control panel (at a pressure of 6 \si{\mega\pascal}).
A custom endoscope camera (1280$\times$720 pixel, 30 FPS, 120\textdegree\ view angle, and 10$\sim$120mm depth of field) is integrated at the end of the continuum manipulator to capture a clear image of the blade surface, and its cables can be hidden in the hollow backbone.

% Because the bending strength of the backbone affects the bending performance of the continuum manipulator, and an overly elastic backbone will cause the continuum manipulator to sag naturally, the use of Ni-Ti hollow tubes can ensure the bending strength and high toughness of the backbone while also achieving a certain degree of lightweight.
% On the basis of this, the guide disks are manufactured using a light-curing printer and resin, effectively reducing weight compared to metal materials.
% For research \cite{torontoiros}, \cite{toronto} similar to ours, our lighter design are crucial for enhancing the length-to-diameter ratio of the continuum manipulator.

% Given the requirement for the tendon to generate significant torque when the backbone bends, with the tendon secured at both ends by a knot, it is imperative for the tendon to possess exceptional tensile and shear strength.
% Therefore, ultra-high-molecular-weight polyethylene fiber is used to manufacture the braided wire, whose tensile strength is fivefold higher than that of nylon, showcasing superior wear resistance. At the same time, the torque of the stepper motor actuating the tendon reaches 1.5 \si{\newton\cdot\meter}.

\begin{figure}[!b]
    \centering
    \includegraphics[width=0.8\linewidth]{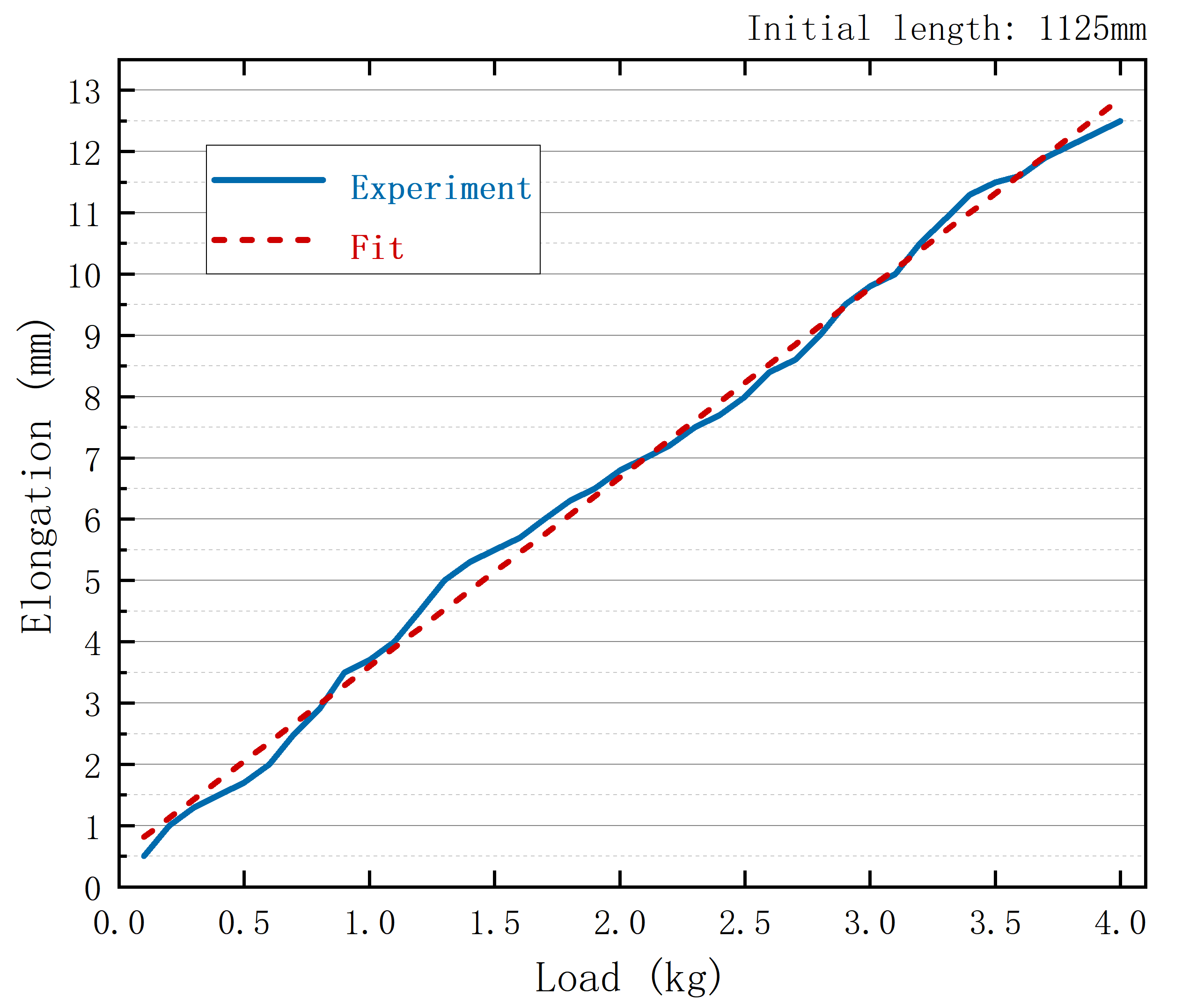}
    \caption{Tension test of a single tendon.}
    \label{fig:tension test}
\end{figure}

% Since the tension of the tendon can be as high as 65 \si{\newton} in the ultimate bending state of the single section backbone, the range and accuracy of the force sensor should be considered, and the experimental test of the tendon's elongation is required.
% Given that the tendon tension can reach up to 65 \si{\newton} in the ultimate bending state of the single-section backbone, it is crucial to consider the range and accuracy of the force sensor. Additionally, conducting experimental tests to measure tendon elongation is essential.
% The tensile test of the tendon with an initial length of 1125 \si{\milli\meter} shows the relationship between the static load and the elongation, as shown in Fig. \ref{fig:tension test}.
% The tendon has a mere elongation rate of 1.408\% under a 4 \si{\kilo\gram} load, demonstrating the feasibility of simplified kinematic model in III.

% To capture a clear image of the blade surface, the continuum manipulator's tip is equipped with a customized endoscope camera (1280$\times$720 pixel, 30 FPS, 120\textdegree\ view angle, and 10$\sim$120mm depth of field), whose cables can be hidden in the hollow backbone.
% During the backbone's pushing and pulling movement, the fixation of the backbone by the pneumatic grippers ensures the effectiveness of the continuum manipulator's shape change.
% To maintain the pneumatic grippers' clamping force, all of them are connected to a pneumatic control plate, which controls the pressure (6 \si{\mega\pascal}) of the gas circuit.

Even though the performance of the PPCM depends on the physical limits of the tendons, our testing of the tendons is limited to typical operating conditions of the PPCM.
The tendon, initially 1125 \si{\milli\meter} in length, shows an elongation rate of only 1.408\% under a 4 \si{\kilo\gram} load, as illustrated in Fig. \ref{fig:tension test} depicting the relationship between static load and elongation.
This not only directly aids in the selection of sensor range and accuracy but also indirectly validates the feasibility of the simplified kinematic model in III.
It is important to note that these values are believed to be below the performance limits of the tendon, as the PPCM can effectively operate with tendon tensions up to 65 \si{\newton} in practical usage scenarios.
Furthermore, permanent deformation of the backbone and tendon breakage are potential issues during PPCM usage.
This is due to the tendons being fixed to the drive shaft, experiencing maximum stress, and being more susceptible to cutting, ultimately resulting in uncontrolled bending of the backbone.
Therefore, regular replacement of severely worn tendons is necessary, along with checking pneumatic pressure to ensure effective movement of the backbone.
\section{KINEMATICS ANALYSIS AND DECOUPLING CONTROL}

\subsection{Methodology Overview}

Based on the piecewise constant curvature (PCC) assumption, the shape of the backbone of each section in the space can be described as a planar arc, which enables the kinematics to be decomposed into two sub-mappings between the three spaces \cite{pcc}. 
By introducing the configuration space between the joint space and the task space of the rigid manipulator, a continuum manipulator, theoretically possessing infinite degrees of freedom, can be fully characterized using a finite number of parameters.

Compared with the Cosserat rod \cite{cosserat1, cosserat2}, Kirchhoff rod \cite{kirchhoff}, and Euler-Bernoulli beam theories \cite{beam1, beam2}, which provide an exact continuous formulation in infinite-dimensional configuration space, in cases where external factors \cite{load} (such as self-weight, load, etc.) and internal factors \cite{friction} (such as tendon hysteresis, friction, etc.) are less influential, the application of a simplified kinematic model has a distinct advantage due to its low computational complexity \cite{control}.
By designing novel control algorithms, it is feasible to relatively improve control performance. In addition, the simplified kinematic model should be developed in a steady state, and path coupling among sections in tendon-driven mechanisms is a fundamental difficulty \cite{coupling}.

In the case of aeroengine inspection, due to the PPCM's difficulty in obtaining prior knowledge of various types of aeroengines' internal structural environments in advance, autonomous motion planning in task space is not possible.
Therefore, teleoperation is an effective method for increasing the versatility of the PPCM in various scenarios, as explained in \cite{intuitiveteleoperation}.
Furthermore, this approach is consistent with the inspectors' practice of using industrial endoscopes, which are more intuitive and user-friendly.
As a result, robot-independent mapping is no longer required, and the robot-specific mapping between configuration space and joint space is the first thing we must investigate.

\begin{figure}[!b]
    \centering
    \includegraphics[width=0.95\linewidth]{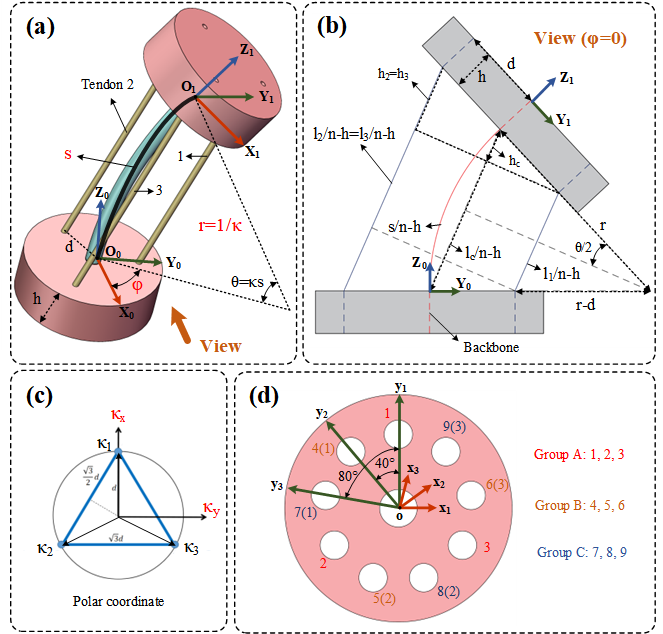}
    \caption{Schematic for the kinematics. a) Definition of coordinate system and configuration space parameters; b) Geometric relationship between tendons and backbone in one segment; c) Viewing curvatures of single tendon as vector quantities to calculate the formulation of tendons' length and backbone parameters; d) Rotate the coordinate system around the z-axis to calculate the length of all tendons passing through a single section.}
    \label{fig:kinematics}
\end{figure}

\subsection{Single Section Kinematics}

To intuitively control the shape of a continuum manipulator, it is essential to derive the Jacobian formulation from joint space (actuator space) to configuration space.
However, the multisection structure discussed in this paper introduces additional complexity to this mapping because the tendons that actuate the distal sections pass through the proximal sections and affect their shape, necessitating a decoupling method.
In consideration of the fixed arc length and curvature of the backbone, along with the continuous variation in planar angle during motion, this section implements a decoupling method based on torsion-free mechanical construction, which necessitates only straightforward coordinate transformations.

Since each section is divided into multiple segments by several guide disks, under the same posture of the continuum manipulator, the relative positions of the backbone and tendons in each segment are identical.
Therefore, by analyzing the geometric relationships of a single segment, it is possible to directly derive the mapping formulation of a single section.
As shown in Fig. \ref{fig:kinematics}(a), frame ${\sum_i}(O_i-X_iY_iY_i)(i=0,1,...,n)$ is fixed in $i^{th}$ guide disk's center, $Y_i$-axis intersects the same tendon (No.1), and $Z_i$-axis is perpendicular to $i^{th}$ guide disk (frame ${\sum_0}$ is the base, only $1^{th}$ segment is shown).
Assuming a specific bending posture where two tendons are of equal length, a series of similar triangle relationships can be employed to establish the current relationship between tendon lengths $(l_1,l_1,l_3)$ and backbone parameters $(\kappa_i,\phi_i,s_i)(i=1,2,3)$, as shown in Fig. \ref{fig:kinematics}(b).
Considering the existence of three such postures and the periodicity of planar angles, the three sets of backbone parameters obtained can be considered as vectors in polar coordinates, as shown in Fig. \ref{fig:kinematics}(c).
By utilizing basic transformation matrices, the parameters of the backbone for any posture can be obtained and represented as

\begin{equation}
\label{eq:forward}
\left\{
\begin{aligned}
    & \kappa_\textrm{single} = \frac{2l_m}{dl_c} \\
    & \phi_\textrm{single} = \arctan{\frac{l_3+l_2-2l_1}{\sqrt{3}(l_2-l_3)}} \\
    & s_\textrm{single} = \frac{ndl_c}{l_m}\arcsin{\frac{l_m}{3nd}}
\end{aligned}
\right.
\end{equation}
where $l_m = \sqrt{l_1^2+l_2^2+l_3^2-l_1l_2-l_2l_3-l_1l_3}$, and $l_c = l_1+l_2+l_3-3nh$. $n$, $d$, and $h$ represent the number of segments, spacing between the tendon and backbone, and height of the guide disk, respectively. Meanwhile, the inverse kinematics from backbone parameters to tendon lengths can be derived using the similar geometric relationships, represented as

\begin{equation}
\label{eq:inverse}
\left\{
\begin{aligned}
    & l_1 = \frac{K(1-d\kappa_\textrm{single}\sin{\phi_\textrm{single}})}{\kappa_\textrm{single}} \\
    & l_2 = \frac{K(1+d\kappa_\textrm{single}\sin{(\phi_\textrm{single}+60^\circ)})}{\kappa_\textrm{single}} \\
    & l_3 = \frac{K(1-d\kappa_\textrm{single}\sin{(\phi_\textrm{single}-30^\circ)})}{\kappa_\textrm{single}}
\end{aligned}
\right.
\end{equation}
where $K = 2n\sin{(\kappa_\textrm{single}(s_\textrm{single}-nh)/2n)}$.

\subsection{Decoupling Method}

For single section, forward and inverse kinematics can be writen as $\mathbf{C}_\textrm{single} = \mathbf{f}(\mathbf{l}_\textrm{single})$ and $\mathbf{l}_\textrm{single} = \mathbf{f}^{-1}(\mathbf{C}_\textrm{single})$, respectively, whose mapping formulation $f$ and $f^{-1}$ have been derived in Subsection B.
For the multisection design discussed in this paper, it is essential to provide specific clarification regarding the configuration space and actuator space in order to apply these kinematic mappings effectively.

The actuator space parameters of the PPCM, denoted as $L_i(i=1,2,...,9)$, can be decomposed into the sum of three single section parameters: $L_i=L_i^\textrm{in}+L_i^\textrm{mid}+L_i^\textrm{out}$, which specifies that the tendon that does not pass through the single section has $0$ length in it.
As shown in Fig. \ref{fig:kinematics}(d), by dividing tendons into three groups, actuator space parameters can be expressed as

\begin{equation}
\label{eq:actuatorparam}
\mathbf{L} = \begin{bmatrix}
    \mathbf{L}_\textrm{A}^\textrm{in} \\
    \mathbf{L}_\textrm{B}^\textrm{in} \\
    \mathbf{L}_\textrm{C}^\textrm{in}
    \end{bmatrix}
    + \begin{bmatrix} \mathbf{L}_\textrm{A}^\textrm{mid} \\
    \mathbf{L}_\textrm{B}^\textrm{mid} \\
    \mathbf{0}
    \end{bmatrix}
    + \begin{bmatrix} \mathbf{L}_\textrm{A}^\textrm{out} \\
    \mathbf{0} \\
    \mathbf{0}
    \end{bmatrix},
\end{equation}
where $\mathbf{L}_\textrm{g}^\textrm{s}(\textrm{g}\in\{\textrm{A},\textrm{B},\textrm{C}\}, \textrm{s}\in\{\textrm{in},\textrm{mid},\textrm{out}\})$ is equivalent to $\mathbf{l}_\textrm{single}$.

Note that $\mathbf{L}_\textrm{A}^\textrm{out}$, $\mathbf{L}_\textrm{B}^\textrm{mid}$, and $\mathbf{L}_\textrm{C}^\textrm{in}$ directly determine the backbone's shape in the proximal, middle, and distal sections, respectively; therefore, they can be calculated using the inverse kinematics mapping $\mathbf{f}^{-1}$.
Considering $\mathbf{L}_\textrm{A}^\textrm{in}$, $\mathbf{L}_\textrm{A}^\textrm{mid}$, and $\mathbf{L}_\textrm{B}^\textrm{in}$ are passively adapted to the backbone, to apply the derived mapping directly, we can rotate the frame about the $Z$-axis so that the $Y$-axis intersects with the other tendons.
Because rotating the frame only affects the planar angle $\phi$, while the arc length $s$ and curvature $\kappa$ of the backbone remain unchanged, the tendon lengths derived from the new backbone parameters $(\kappa,\phi_\textrm{new},s)$ in new frame correspond to other three tendons.
For example, Fig. \ref{fig:kinematics}(d), the lengths of tendon 4, 5, and 6 in this single section can be derived using the inverse kinematics of the backbone parameters in the $x_2Oy_2$ frame rotated counterclockwise by $40^\circ$ around the $z$-axis, where the $y_2$-axis passes through tendon 4.
Based on this, the mapping between the PPCM's actuator space and configuration space $\mathbf{C}=[\mathbf{C}_\textrm{in}, \mathbf{C}_\textrm{mid},\mathbf{C}_\textrm{out}]^T$ can be expressed as

\begin{equation}
\label{eq:actuatorparamnew}
\mathbf{L} = \begin{bmatrix}
    \mathbf{g}_1(\mathbf{C}_\textrm{in}) \\
    \mathbf{g}_2(\mathbf{C}_\textrm{in}) \\
    \mathbf{f}^{-1}(\mathbf{C}_\textrm{in})
    \end{bmatrix}
    + \begin{bmatrix} \mathbf{g}_2(\mathbf{C}_\textrm{mid}) \\
    \mathbf{f}^{-1}(\mathbf{C}_\textrm{mid}) \\
    \mathbf{0}
    \end{bmatrix}
    + \begin{bmatrix} \mathbf{f}^{-1}(\mathbf{C}_\textrm{out}) \\
    \mathbf{0} \\
    \mathbf{0}
    \end{bmatrix},
\end{equation}
where $\mathbf{g}_2(\kappa,\phi,s)=\mathbf{f}^{-1}(\kappa,\phi+40^\circ,s)$ and $\mathbf{g}_1(\kappa,\phi,s)=\mathbf{g}_2(\kappa,\phi+40^\circ,s)$. Therefore, the inverse Jacobian formulation of the PPCM from the configuration space to actuator space can be obtained by converting (\ref{eq:actuatorparamnew}) into differential form, expressed as

\begin{equation}
\label{eq:jacobian}
\mathbf{\dot{L}} = \begin{bmatrix}
    \dot{\mathbf{g}_1}(\mathbf{C}_\textrm{in}) &\dot{\mathbf{g}_2}(\mathbf{C}_\textrm{mid}) &\mathbf{J}^{-1}(\mathbf{C}_\textrm{out}) \\
    \dot{\mathbf{g}_2}(\mathbf{C}_\textrm{in}) &\mathbf{J}^{-1}(\mathbf{C}_\textrm{mid}) &\mathbf{0} \\
    \mathbf{J}^{-1}(\mathbf{C}_\textrm{in}) &\mathbf{0} &\mathbf{0}
    \end{bmatrix}
    \mathbf{\dot{C}},
\end{equation}
where $\dot{\mathbf{l}}_\textrm{single} = \mathbf{J}^{-1}(\dot{\mathbf{C}}_\textrm{single})$.

\subsection{Closed-loop Control}

Given the slackness of tendons under open-loop control \cite{taskspace}, it is essential to consider the tension of the tendons to ensure the effectiveness of the mapping derived in Subsection C.
The advantages of the PID control algorithm have been validated in the continuum manipulator \cite{pid}.
Incorporating the error in tendon tension as a feedforward loop effectively ensures the tautness of the tendons without requiring a complex model.
The entire control procedure is elucidated in Algorithm \ref{alg:control}.

\begin{breakablealgorithm}
\renewcommand{\algorithmicrequire}{\textbf{Input:}}
\renewcommand{\algorithmicensure}{\textbf{Output:}}
\caption{Decoupling and Closed-loop Control}
\label{alg:control}
\begin{algorithmic}[1]
        \REQUIRE Desired velocity of 3 single sections in configuration space $\mathbf{\dot{C}}=[\mathbf{\dot{C}}_\textrm{in}, \mathbf{\dot{C}}_\textrm{mid},\mathbf{\dot{C}}_\textrm{out}]^T$ and initial position of 9 tendons in actuator space (length) $\mathbf{L}=[L_1, L_2, ..., L_9]^T$.
        
        \ENSURE Velocity of 9 tendons in actuator space $\mathbf{\dot{L}}$.
        
            \STATE Give number of segments in 3 single sections $n_\textrm{in}$, $n_\textrm{mid}$, and $n_\textrm{out}$
            
            \STATE Give spacing between tendon and backbone $d$
            
            \STATE Set length range $s_\textrm{in} \in (s_\textrm{in}^{min}, s_\textrm{in}^{max})$, $s_\textrm{mid} \in (s_\textrm{mid}^{min}, s_\textrm{mid}^{max})$, $s_\textrm{out} \in (s_\textrm{out}^{min}, s_\textrm{out}^{max})$
            
            \STATE Set curvature range $\kappa_\textrm{in} \in (\kappa_\textrm{in}^{min}, \kappa_\textrm{in}^{max})$, $\kappa_\textrm{mid} \in (\kappa_\textrm{mid}^{min}, \kappa_\textrm{mid}^{max})$, $\kappa_\textrm{out} \in (\kappa_\textrm{out}^{min}, \kappa_\textrm{out}^{max})$
            
            \STATE Initialize controller parameters $\mathbf{K}_p$, $\mathbf{K}_i$, $\mathbf{K}_d$, $\mathbf{F}_\textrm{ref}$

            \STATE Integral $\mathbf{F}_\textrm{i}=0$ and Differential $\mathbf{F}_\textrm{d}=0$ of tension error
            
            \WHILE{$t < T$ and $\kappa_\textrm{sec},s_\textrm{sec}(\textrm{sec}\in\{\textrm{in},\textrm{mid},\textrm{out}\}))$ are in range}

                \STATE $\mathbf{C}_\textrm{in}=\mathbf{f}(L_7,L_8,L_9)$, then $L_5^{\textrm{in}},L_6^{\textrm{in}},L_7^{\textrm{in}}=\mathbf{g}_2(\mathbf{C}_\textrm{in})$, and $L_1^{\textrm{in}},L_2^{\textrm{in}},L_3^{\textrm{in}}=\mathbf{g}_1(\mathbf{C}_\textrm{in})$

                \STATE $\mathbf{C}_\textrm{mid}=\mathbf{f}(L_5-L_5^{\textrm{in}},L_6-L_6^{\textrm{in}},L_7-L_7^{\textrm{in}})$, then $L_1^{\textrm{mid}},L_2^{\textrm{mid}},L_3^{\textrm{mid}}=\mathbf{g}_2(\mathbf{C}_\textrm{mid})$

                \STATE $\mathbf{C}_\textrm{out}=\mathbf{f}(L_1-L_1^{\textrm{in}}-L_1^{\textrm{mid}},L_2-L_2^{\textrm{in}}-L_2^{\textrm{mid}},L_3-L_3^{\textrm{in}}-L_3^{\textrm{mid}})$

                \STATE Inverse kinematics $\mathbf{\dot{L}}_\textrm{IK}=\mathbf{J}_\textrm{multi}^{-1}\mathbf{\dot{C}}$

                \STATE PID compensation $\mathbf{\dot{L}}_\textrm{c}=\mathbf{K}_p(\mathbf{F}_\textrm{t}-\mathbf{F}_\textrm{ref})+\mathbf{K}_i\mathbf{F}_\textrm{i}+\mathbf{K}_d\mathbf{F}_\textrm{d}$
                
                \STATE Update $\mathbf{F}_\textrm{i}$ and $\mathbf{F}_\textrm{d}$

                \STATE $\mathbf{\dot{L}}=\mathbf{\dot{L}}_\textrm{IK}+\mathbf{\dot{L}}_\textrm{c}$

            \ENDWHILE
\end{algorithmic}
\end{breakablealgorithm}

\section{EXPERIMENT AND RESULTS}

\begin{figure}[!h]
    \centering
    \includegraphics[width=0.95\linewidth]{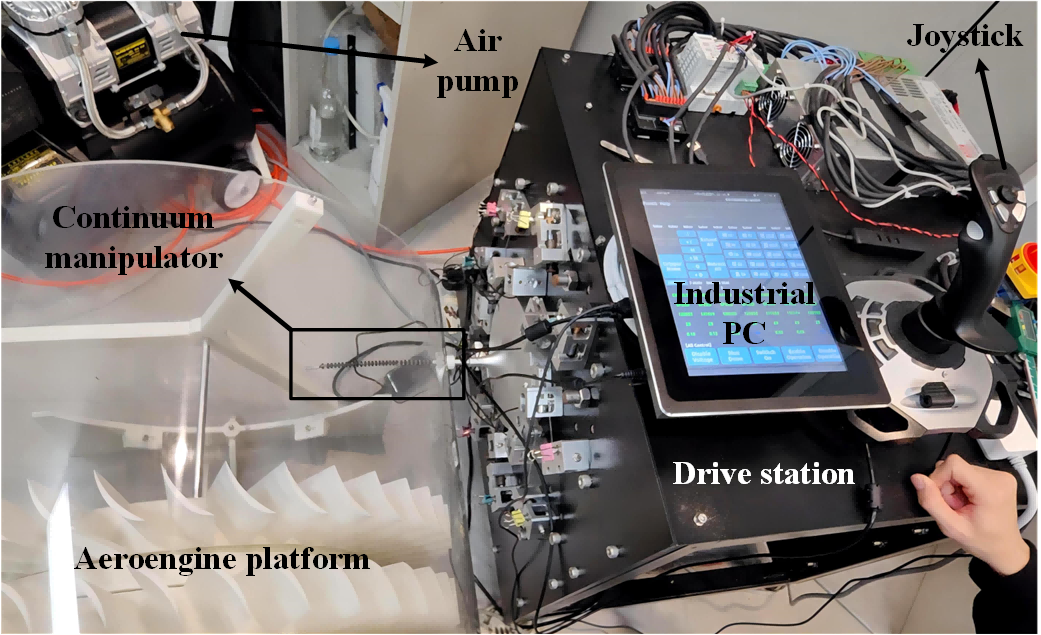}
    \caption{The PPCM prototype and a 3D-printed aeroengine platform. Teleoperation is achieved through an external joystick that is connected to an industrial PC.}
    \label{fig:prototype}
\end{figure}

An experimental layout is presented in Fig. \ref{fig:prototype}.
Note that after each power-up of the PPCM, it needs to be calibrated to ensure consistency in the initial posture (with the shortest length and zero curvature of all sections) of the continuum manipulator, which can be achieved by implementing constant force control on the tendons.
Open all pneumatic grippers to allow axial movement of the three backbones, with all tendons in a relaxed state (tension approximately zero).
Until each tendon reaches its desired tension, it will be continuously stretched, causing the corresponding backbone affected by this tendon to continuously deform (shorten or/and bend) along with the tendon's movement.
This deformation can be manually controlled to maintain all backbones in a horizontal posture through hand guidance.
Simultaneously, tendons adaptively move (tighten or loosen) in response to changes in tension until the continuum manipulator reaches a steady state, at which point measuring the length of each section yields the initial kinematics parameters.

\subsection{Validation of kinematics and Decoupling Method}

\begin{figure}[!tbh]
    \centering
    \includegraphics[width=0.9\linewidth]{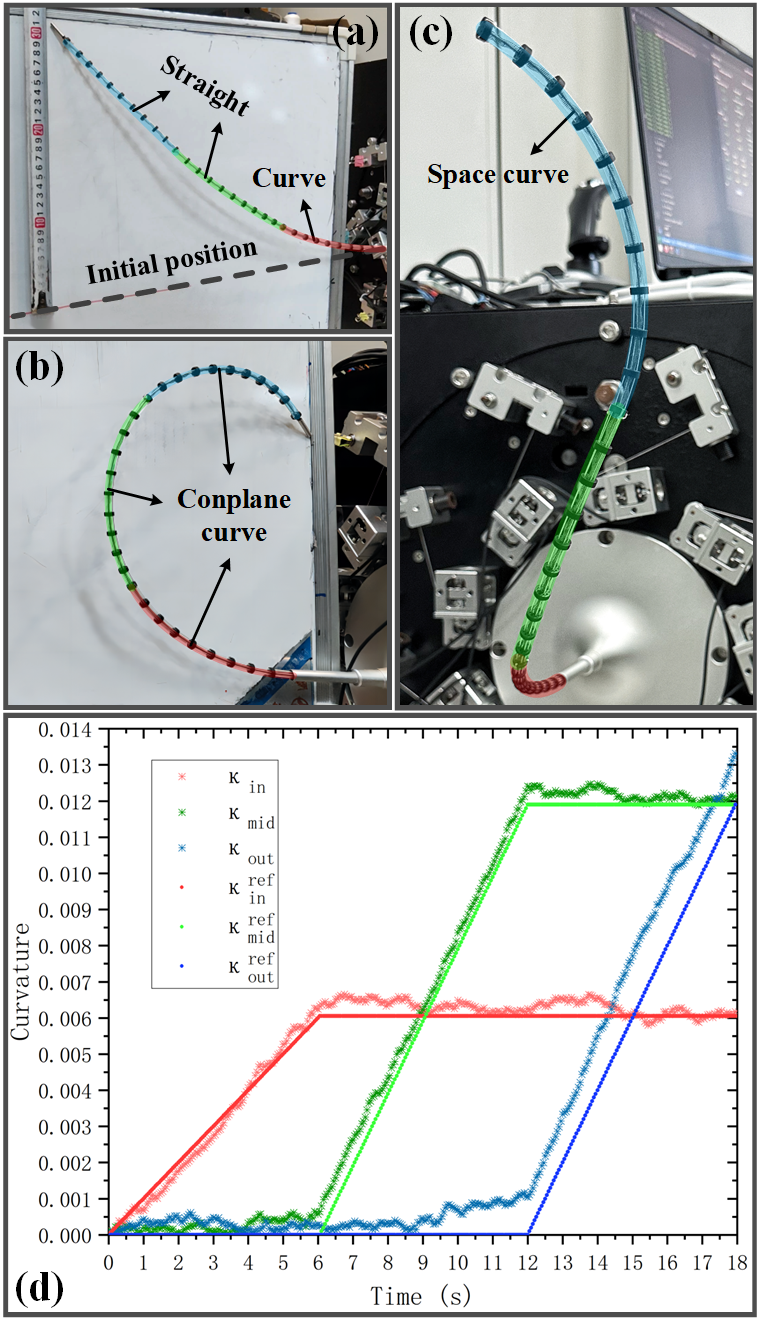}
    \caption{
    Results of the validation of the decoupling method of kinematics.
    Sub-figure (a) shows the proximal section (red curve) being bent first does not affect the straight state of the middle section (green line) and distal section (blue line).
    Sub-figure (b) shows that bending the middle section (green curve) and distal section (blue curve) in sequence also does not alter the curvature of the proximal section and can cause a significant rotation of the end effector in the plane (greater than 180\textdegree).
    Sub-figure (c) shows that rotating the plane of the three sections allows the continuum manipulator to form an arbitrary spatial curve.
    Sub-figure (d) illustrates the changes in curvature for each section throughout the aforementioned entire motion process.
    }
    \label{fig:validation}
\end{figure}

The decoupling method of kinematics is validated by sequentially applying velocity control to the configuration space of each section and observing the resulting changes in the posture of other sections. As shown in Fig. \ref{fig:validation}, 
the experimental results indicate that there is a certain deviation between the curvature of each section and the expected value at most moments, primarily compensating for tendon slackness during movement. However, the overall motion trend remains consistent, suggesting that there is no mutual effect among sections, confirming decoupling.

\subsection{Inspection Experiment}

\begin{figure*}[!t]
    \centering
    \includegraphics[width=0.9\textwidth]{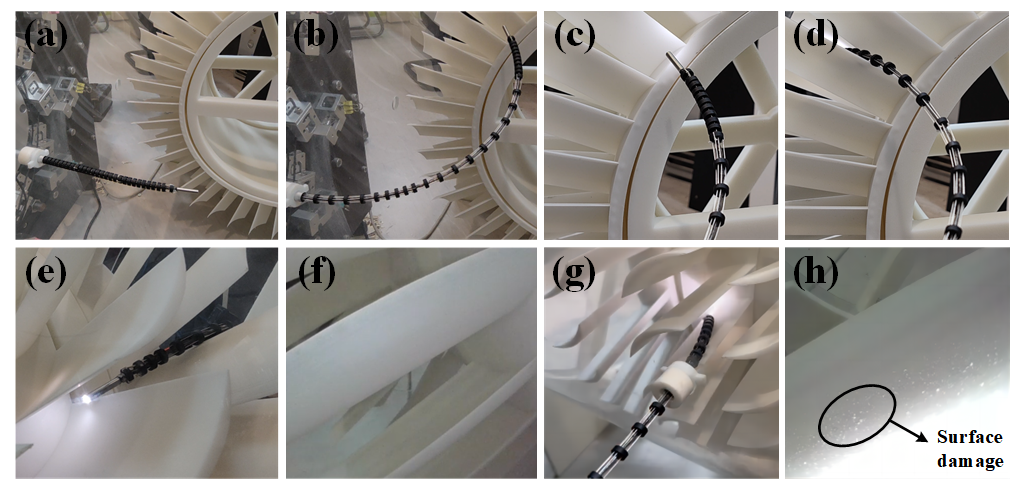}
    \caption{Inspection experiment of the aeroengine blades. Sub-figures (a) to (e) show the performance of PPCM in a turbine inspection scenario.
    By bending and extending the proximal and middle sections, the continuum manipulator moves away from its initial position, positioning the endoscope camera closer to the turbine, while the distal section remains contracted and straight.
    Bending and extending the distal section allows the continuum manipulator to pass through the gap and enter the next stage, the image captured by the endoscope camera is shown in sub-figure (f).
    Sub-figure (g) shows the continuum manipulator entering the inspection port in the middle of the compressor to inspect its blades.
    Image of the blade surface in sub-figure (h) can directly validate the feasibility of the continuum manipulator in aeroengine blade inspection.}
    \label{fig:inspection experiment}
\end{figure*}

The PPCM showcases superior performance in practical inspection experiments with its push-pull multisection design.
These experiments were conducted on a significantly smaller experimental platform compared to previous work, and the motion process is illustrated in Fig. \ref{fig:inspection experiment}. 
The PPCM achieves both extensive range of motion and precise maneuvering of its end effector, which is a capability beyond conventional continuum manipulator mechanisms. 
Similar to mobile manipulation, the proximal and middle sections can be seen as a platform providing flexible spatial movement for the distal section.
As the proximal section approaches the blades closely, while maintaining the shape of the continuum manipulator unchanged, strategies are devised to guide the detection probe through narrow gaps between the blades.
The images of the blade surfaces captured by the endoscopic camera confirm the feasibility of the PPCM in aerospace engine blade detection tasks and demonstrate its adaptability across various task scenarios.

% The process and results of the inspection experiment are shown in Fig. \ref{fig:inspection experiment}.
% In the turbine scenario, the proximal and middle sections are used to move the endoscope camera over a wide range, while the distal section can be regarded as a sub-manipulator, controlling the position and attitude of the PPCM's tip finely, allowing it to traverse the gap and inspect multistage blades.
% On the contrary, the inspection port used to inspect the compressor blades is located in the middle, and the space for the PPCM's movement is limited, making it more significant for the planning and control of the distal section.
% Through the image of blade surface captured by the endoscope camera at close range, the feasibility of the PPCM in the aeroengine blade inspection task has been verified, and it has the adaptability of several different task scenarios.

\section{CONCLUSIONS AND FUTURE WORK}

In this paper, we have proposed an ultra-slender (length-to-diameter ratio \textgreater\ 60) push-pull multisection continuum manipulator for the inspection of compressor and turbine blades in aeroengines through external inspection ports.
To improve the capability to traverse the confined space between blades, especially in small engines, we have developed a push-pull multisection structure combined with a tendon-driven mechanism, allowing the PPCM to inspect multistage blades through a single inspection port.
A kinematic model for the PPCM has been derived, which simplifies the submapping transformation and Jacobian formulation for multiple sections and introduces a decoupling method. 
We successfully constructed a prototype with an experimental platform, confirmed the effectiveness of kinematics and the decoupling method, implemented closed-loop control, and carried out blade inspection tasks through teleoperation.

In future work, we intend to optimize the structure design to enhance portability. Subsequently, we aim to integrate it onto a specific mobile platform, such as an in-pipe robot, to access the aircraft's jet nozzle and inspect the hidden engine without requiring reserved inspection ports.

% \addtolength{\textheight}{-12cm} 

\bibliographystyle{unsrt}
\bibliography{reference}

\end{document}